# Tokenization Strategies for Low-Resource Agglutinative Languages in Word2Vec: Case Study on Turkish and Finnish

Jinfan Frank Hu
Phillips Academy
Andover, Massachusetts, United States
jinfanfrank@gmail.com

*Abstract— Tokenization plays a critical role in processing agglutinative languages, where a single word can encode multiple morphemes carrying syntactic and semantic information. This study evaluates the impact of various tokenization strategies—word-level, character-level, n-gram, and Byte Pair Encoding (BPE)—on the quality of static word embeddings generated by Word2Vec for Turkish and Finnish. Using a 10,000-article Wikipedia corpus, we trained models under low-resource conditions and evaluated them on a Named Entity Recognition (NER) task. Despite the theoretical appeal of subword segmentation, word-level tokenization consistently outperformed all alternatives across all tokenization strategies tested. These findings suggest that in agglutinative, low-resource contexts, preserving boundaries via word-level tokenization may yield better embedding performance than complex statistical methods. This has practical implications for developing NLP pipelines for under-resourced languages where annotated data and computing power are limited.*

*Keywords— Preprocessing, Tokenization, Agglutinative Languages, Word2Vec, Low-Resource Natural Language Processing*

## I. Introduction

In recent years, large language models (LLMs) have revolutionized natural language processing by learning rich patterns from massive text corpora. A foundational step in training these models is *tokenization*: the process of breaking text into discrete units, or tokens, that serve as the model's inputs. Once tokenized, each unit is mapped to a vector, forming the building blocks of the model's understanding of language. While modern LLMs use dynamic, contextual embeddings that shift with surrounding context, earlier approaches like Word2Vec rely on *static embeddings*, where each token is assigned a fixed vector. Though less powerful overall, these static models serve as a valuable tool for studying the isolated effects of different tokenization strategies.

Tokenization becomes especially critical in languages with complex morphology. Agglutinative languages such as Turkish and Finnish construct words by stringing together affixes, often encoding meanings that would require full phrases in English. For example, the Turkish word *evlerimizde* translates to "in our homes," blending a root "ev" with plural "ler", possessive "imiz", and locative "de" suffixes. In such cases, naive word-level tokenization can lead to sparsity (rare words containing no meaningful data), vocabulary bloat (redundancies in generating embeddings), and weakened generalization (the ability to adapt to unseen data) in NLP systems.

To address this, researchers often turn to subword segmentation strategies such as character-level tokenization, n-grams, or Byte Pair Encoding (BPE), which aim to break words into smaller, reusable units. These techniques promise to reduce sparsity and improve generalization, especially for morphologically rich or low-resource languages. However, the actual performance tradeoffs of these methods, especially on small or mid-sized corpora, remain underexplored.

In this study, we investigated how different tokenization strategies affected the quality of static embeddings in agglutinative languages. We trained Word2Vec models on Turkish and Finnish Wikipedia corpora using four tokenization methods: word-level, character-level, n-grams, and BPE. To evaluate downstream utility, we tested each model on Named Entity Recognition (NER) tasks. While Turkish and Finnish are relatively well-resourced, they offered a valuable window into the broader challenges of processing morphologically rich languages with limited data.

Surprisingly, our findings show that simple word-level tokenization often outperforms more modern subword methods, especially on smaller corpora. These results suggest that linguistic structure, rather than purely statistical segmentation, may offer greater advantages in certain contexts. For practitioners working with under-resourced languages, this insight could inform the design of more efficient, linguistically aligned NLP pipelines, even challenging the subword-centric strategies employed by many modern multilingual LLMs.

## II. Related Work

Tokenization is one of the earliest and most extensively studied tasks in natural language processing [1]. In agglutinative languages such as Finnish, Estonian, and Turkish, complex morphological structures present challenges for standard speech-to-text and text segmentation tools. Traditional word-level tokenization often leads to out-of-vocabulary (OOV) errors due to the theoretically infinite ways of constructing vocabulary in such languages [2]. To address this, rule-based morphological analysis and statistical segmentation methods have been explored, both demonstrating improvements over whitespace-based approaches [3], [1].

Unsupervised subword tokenization techniques, such as BPE, have also shown promise in reducing OOV rates, though their effectiveness can vary depending on the downstream task and language morphology [4]. These approaches are particularly relevant for low-resource languages, where limited data availability exacerbates common challenges such as sparsity and generalization in NLP systems [5].

Prior work has highlighted the broader implications of improving NLP for low-resource agglutinative languages, including digital inclusion, cultural preservation, and the expansion of language technology to underrepresented populations [5]. Comparative studies involving Quechua and Finnish have demonstrated the feasibility of cross-linguistic tokenization strategies, but few works have evaluated how basic preprocessing methods—when paired with classic embedding models like Word2Vec—perform on concrete evaluation tasks such as Named Entity Recognition (NER) using modest corpora [4].

This paper addresses that gap by systematically evaluating tokenization strategies on Turkish and Finnish using limited Wikipedia corpora, simulating low-resource conditions and assessing performance on NER tasks.

### III. METHODOLOGY

This study uses Wikipedia corpora in Turkish and Finnish to simulate low-resource language conditions. After downloading Wikipedia's native XML dumps, automated welcome messages and associated metadata were removed. A random sample of 10,000 articles was selected per language. This relatively small subset was chosen to reflect the size and quality of available resources for many low-resource agglutinative languages.

The XML files were converted to JSON and then preprocessed using Python. We then used Python to strip formatting artifacts, punctuation, and markup. The resulting corpora, consisting of only characters, digits, and whitespace, were saved as plain text.

Five tokenization strategies were applied: word-level, character-level, bigrams, trigrams, and BPE. BPE merges the most frequent character pairs until a target vocabulary size is reached.

For example, "abcbabcbab" would be merged as follows:

"a" + "b" = "ab" → "c" + "b" = "cb" → "ab" + "cb" = "abcb" and so on, until all characters are merged or an arbitrary vocabulary size is reached.

Five BPE variants were trained with vocabulary sizes of 5,000, 10,000, 25,000, 50,000, and 100,000 subwords. Modern multilingual models usually employ vocabulary sizes of about 100,000 [6] so 50,000 should be enough for a monolingual model, especially if the goal is to avoid segmenting common morphemes. The BPE algorithm followed the implementation from the *YouTokenToMe* library [7].

For non-BPE tokenization strategies, the Hugging Face datasets library was used to preprocess and convert the data into *.arrow* format [8]. Tokenization was implemented using sliding windows or whitespace splitting.

Word embeddings were trained using *gensim's* Word2Vec implementation, with a vector size of 150 [9]. Performance was evaluated using Named Entity Recognition (NER), in which tokens are labeled with entity types such as B-Person, I-Date, or O (outside of any entity). Named Entity Recognition is often used to evaluate Word2Vec and other embedding models because it tests how well the model's embeddings capture semantic and syntactic information.

"B" corresponds to "Beginning," (the first word that corresponds to an entity) while "I" corresponds to "Inside" (words that express a continuation of an entity). For example,

| Barack | Obama | lives | in | Honolulu | . |
|---|---|---|---|---|---|
| B-Person | I-Person | O | O | B-Location | O |

Likewise, an example (with English translation) from the NER sets used is as follows:

| Yalnız | Adam | şarkısıyla | tanınmıştır | . |
|---|---|---|---|---|
| Yalnız | Adam | with the song | became famous | . |
| B-ART | I-ART | O | O | O |

Or more directly translated, "He became famous with the song *Yalnız Adam*."

The Turkish data came from the Turkish NLP Suite, already split into training and test sets in a 90/10 split. For Finnish, a dataset from MetaText.io was randomly divided 90/10 for training and evaluation [10][11]. Notably, the Turkish dataset was significantly larger and more diverse than its Finnish counterpart.

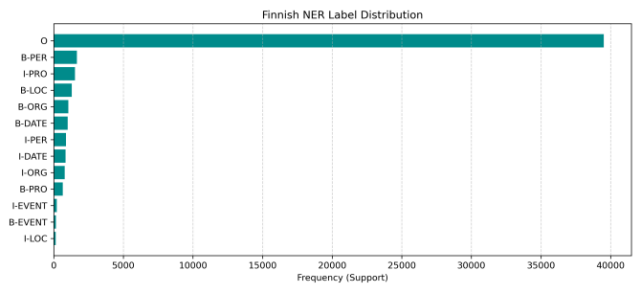

Fig. A. This histogram shows uneven distribution between labels for Finnish, with most notably, "O" serving as a large majority of the labels. This will explain label skew in evaluation.

Classification was performed using logistic regression with the Stochastic Average Gradient with Adaptivity (SAGA) solver from the *sklearn* library [12]. This solver was chosen for its convergence efficiency and suitability for large datasets with limited computational resources. Training was capped at 500 epochs, which was sufficient for convergence (<0.0001 change per epoch), reducing computational strain and reducing the risk of overfitting. Final model evaluations were saved in *.json* format for analysis.

Following initial experiments that revealed implementation issues with BPE tokenization and entity tag propagation across different tokenization schemes, the final experimental setup included five BPE variants with vocabulary sizes of 5,000, 10,000, 25,000, 50,000, and 100,000 subwords. Entity tags were also retrospectively propagated to match the tokenization boundaries of each strategy, ensuring fair comparison across different segmentation approaches.

## IV. RESULTS

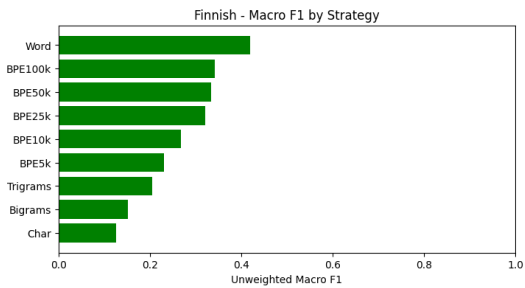

Fig. B. This histogram shows unweighted macro F1 scores (an average of each entity's F1 score, a metric balancing precision and recall) for each tokenization strategy in Finnish.

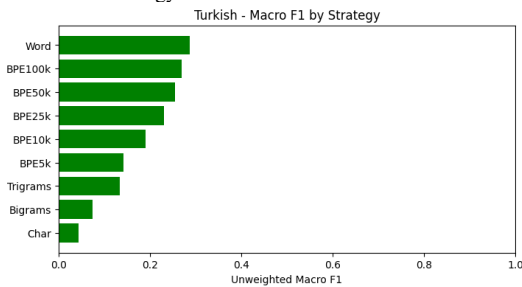

Fig. C. This is a histogram for macro F1 scores for Turkish.

For both languages, Word-level tokenization had the best performance by unweighted macro F1 scores, followed by BPE100k, BPE50k, BPE25k, BPE10k, BPE5k, then Trigrams, Bigrams, and Character.

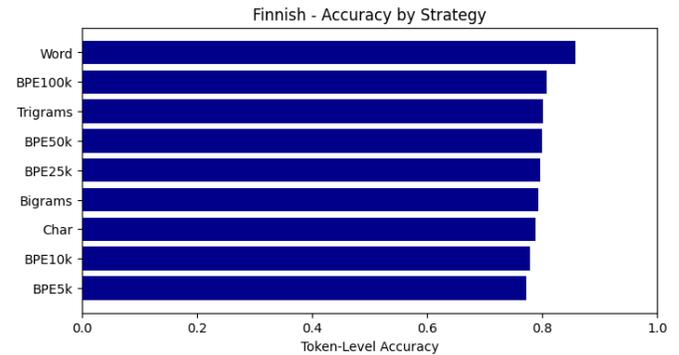

Fig. D. This histogram shows accuracy (correct predictions divided by total predictions) scores for each tokenization strategy in Finnish.

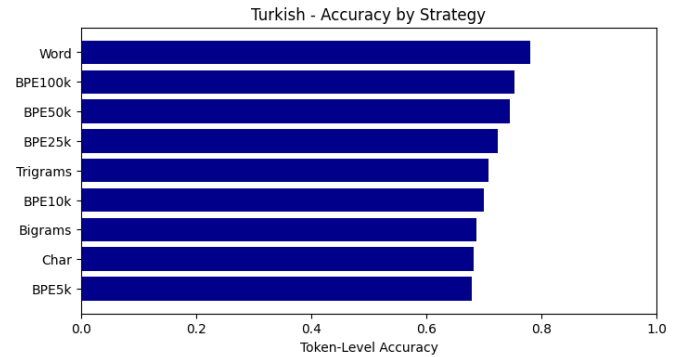

Fig. E. This histogram shows accuracy scores for each tokenization strategy in Turkish.

Accuracy scores for each language varied, with Word-level again taking the top spot, with a mix of BPE and n-gram strategies following, with larger tokenization units generally achieving higher accuracies.

These plots highlight a key distinction between accuracy and macro F1 as evaluation metrics. Accuracy tends to reward models that default to the majority class ("O"), while macro F1 exposes differences in how well each model recognizes minority entity classes. For instance, while accuracy dropped slightly for some subword models due to fewer correct "O" predictions, their macro F1 scores improved significantly, showing better performance on actual named entities.

Notably, word-level tokenization maintained the highest macro F1 across both languages, suggesting that semantic coherence from preserving full words continues to provide a competitive edge. Meanwhile, BPE strategies showed steadily increasing F1 with larger vocabulary sizes, reinforcing the idea that subword segmentations benefit from greater lexical coverage.

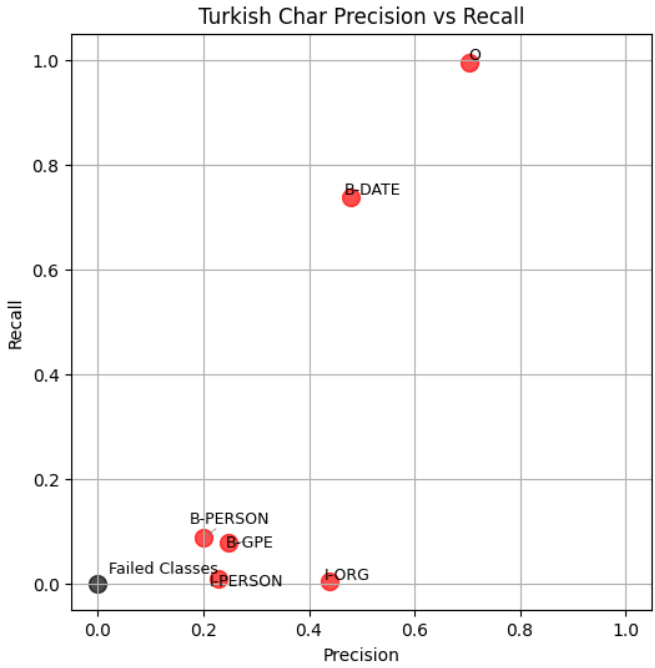

Fig. F. This plot of precision (correct predictions for a class divided by total predictions for said class) against recall (correct predictions for a class divided by total instances for said class) for the Turkish Character-level tokenizer displays a general lack of understanding for most classes, as only 6/39 (~15%) classes had any precision or recall.

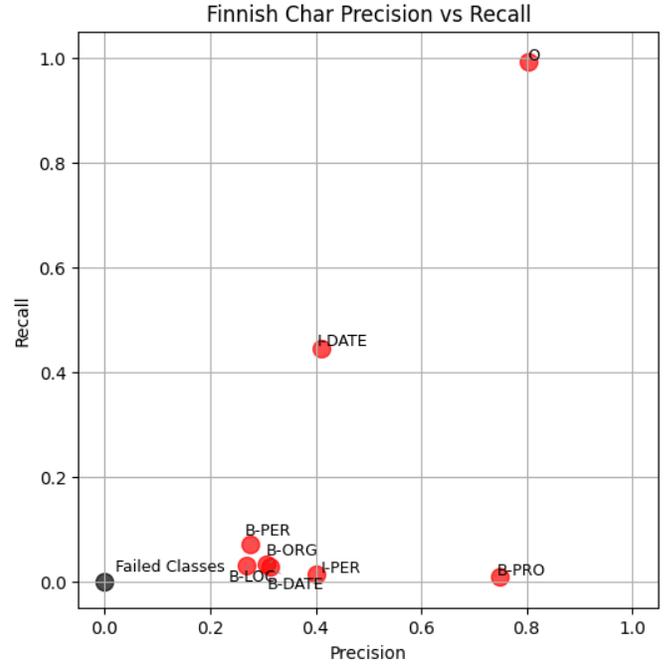

Fig. H. This plot for the Finnish Character-level tokenizer demonstrates poor understanding of the NER set's classes, with 7/13 (~54%) classes with precision or recall.

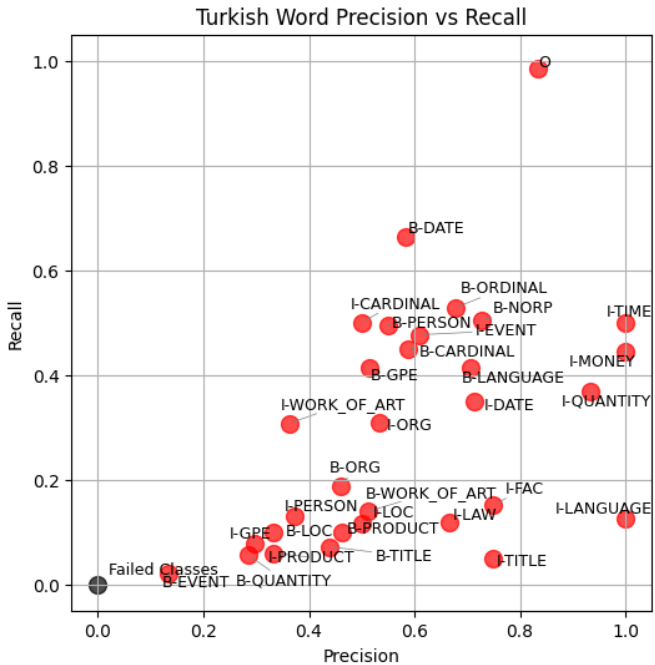

Fig. G. This plot for the Turkish Word-level tokenizer shows a solid understanding of many classes, with some kind of understanding of 28/39 (~72%) classes in the NER set.

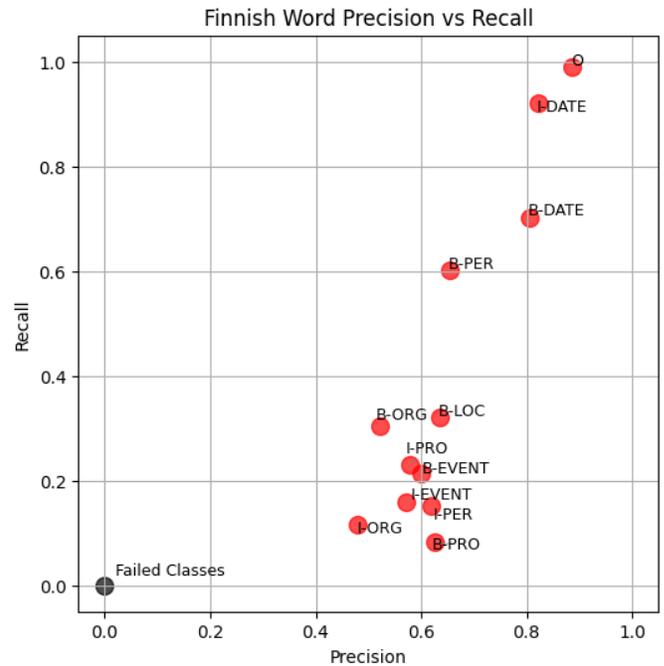

Fig. I. This plot for the Finnish Word-level tokenizer demonstrates a very good understanding of the NER set's classes, with 12/13 (~92%) classes with precision or recall. The only class that failed was "I-LOC", which only had a support of n=5. The next least-supported class was "B-EVENT", with n=14.

Overall, the results reinforce the value of preserving full word boundaries in agglutinative languages when working under low-resource constraints. Word-level tokenization consistently achieved the highest macro F1 scores across both Turkish and Finnish, outperforming more granular segmentation strategies such as characters, n-grams, and even BPE. This held true even after correcting label alignment across tokenization schemes and using morphologically rich corpora. These findings suggest that full-word units, despite their higher vocabulary size and sparsity, may better encode syntactic and semantic structure, particularly in settings where entity recognition requires understanding multi-morpheme compounds.

Subword strategies exhibited more nuanced behavior. While character-level and n-gram approaches suffered from fragmented semantic units and limited precision, BPE models showed marked improvement as vocabulary size increased. The highest-performing BPE model (100,000 vocabulary) approached word-level performance, especially in Finnish, indicating that when BPE units begin approximating full-word morphemes, they become viable alternatives. However, their effectiveness remained sensitive to vocabulary size and data quality, suggesting a trade-off between generalizability and linguistic coherence. Precision-recall plots further highlighted that word-level models were more consistent across entity classes, especially those with moderate to high support, while subword models often struggled with recall on rarer tags. These patterns emphasize the need to align tokenization granularity with linguistic structure and task-specific goals.

## V. Conclusion

This study demonstrates that tokenization strategy plays a critical role in training word embedding models for agglutinative languages. Word-level tokenization consistently outperformed subword approaches on NER tasks, particularly when paired with high-quality, well-structured text. While subword methods such as BPE improved with larger vocabularies, they remained less effective overall on moderate corpora. These findings suggest that, for many low-resource languages, preserving linguistic structure may outweigh the theoretical efficiency of statistical segmentation.

In practical terms, developers building NLP tools for under-resourced or indigenous languages should prioritize word-level tokenization over subword methods when working with limited training data, as our results demonstrate consistent performance advantages. These results also suggest that in agglutinative languages, the semantic coherence of full-word units may outweigh the statistical advantages of subword segmentation, especially in low-resource settings, unless vocabulary sizes become large enough to approximate full-word tokenization.

While this study provides insight into the effectiveness of tokenization strategies for agglutinative languages, it is not without limitations. The corpora for Turkish and Finnish, though moderately sized, do not fully represent the conditions of truly low-resource languages, especially in terms of orthographic variation, dialectal diversity, and noisy or informal data. Additionally, the study focused exclusively on static embeddings and NER tasks, which may not fully capture the range of downstream effects in more complex NLP pipelines. The NER datasets also varied in quality: the Turkish dataset was about ten times larger than Finnish's, with more diverse and varied label categories as well. Therefore, we were unable to make a comparative study of the two languages' performance. Despite this, the overall trends across both languages suggest the robustness of this study's findings.

Future work could investigate whether these patterns persist under contextual embedding architectures such as BERT or ByT5 and assess their generalizability to other morphologically rich languages [6], [14]. These architectures may also compensate for smaller vocabulary sizes in BPE or n-grams models for small n's. And while Turkish and Finnish served as proxies, future studies using annotated NER datasets in truly low-resource languages could validate these findings more directly, helping advance linguistic equity and enhancing representation in digital tools.

By incorporating a more diverse and varied corpus, such as text message conversations and social media forums, this research could bridge the gap between clean, theoretical NER sets and real-world applications [15]. Beyond NER, tasks such as POS tagging, dependency parsing, or even semantic similarity might respond differently to tokenization strategies [16], [17]. Multi-task setups could reveal broader implications for tokenizers in NLP and LLM pipelines.

### Acknowledgment

Jinfan Frank Hu would like to thank Prof. Ganesh Mani of Carnegie Mellon University for supporting him throughout the writing of the paper. He would also like to thank Dhara Mungra for keeping him on track with the paper writing process.

The code and files for this project can be found at https://github.com/jinfanfrankhu/TokenizationResearch